\documentclass[10pt,twocolumn,letterpaper]{article}

\usepackage{cvpr}              

\usepackage{graphicx}
\usepackage{amsmath}
\usepackage{amssymb}
\usepackage{booktabs}
\usepackage{rotating}
\usepackage[accsupp]{axessibility}  

\usepackage[pagebackref,breaklinks,colorlinks]{hyperref}
\usepackage{multirow}

\usepackage[capitalize]{cleveref}
\crefname{section}{Sec.}{Secs.}
\Crefname{section}{Section}{Sections}
\Crefname{table}{Table}{Tables}
\crefname{table}{Tab.}{Tabs.}

\def\cvprPaperID{7239} 
\def\confName{CVPR}
\def\confYear{2022}

\begin{document}

\title{Anomaly Detection via Reverse Distillation from One-Class Embedding}

\author{Hanqiu Deng \hspace{1cm} Xingyu Li\\
Department of Electrical and Computer Engineering, University of Alberta\\
{\tt\small \{hanqiu1,xingyu\}@ualberta.ca}
}
\maketitle

\begin{abstract}

Knowledge distillation (KD) achieves promising results on the challenging problem of unsupervised anomaly detection (AD). The representation discrepancy of anomalies in the teacher-student (T-S) model provides essential evidence for AD. However, using similar or identical architectures to build the teacher and student models in previous studies hinders the diversity of anomalous representations. To tackle this problem, we propose a novel T-S model consisting of a teacher encoder and a student decoder and introduce a simple yet effective "reverse distillation" paradigm accordingly. Instead of receiving raw images directly, the student network takes teacher model's one-class embedding as input and targets to restore the teacher's multi-scale representations. Inherently, knowledge distillation in this study starts from abstract, high-level presentations to low-level features. In addition, we introduce a trainable one-class bottleneck embedding (OCBE) module in our T-S model. The obtained compact embedding effectively preserves essential information on normal patterns, but abandons anomaly perturbations. Extensive experimentation on AD and one-class novelty detection benchmarks shows that our method surpasses SOTA performance, demonstrating our proposed approach's effectiveness and generalizability.
\end{abstract}

\section{Introduction}
\label{sec:intro}
Anomaly detection (AD) refers to identifying and localizing anomalies with limited, even no, prior knowledge of abnormality. The wide applications of AD, such as industrial defect detection \cite{mvtec}, medical out-of-distribution detection \cite{mood}, and video surveillance \cite{stc}, makes it a critical task as well as a spotlight. In the context of unsupervised AD, no prior information on anomalies is available. Instead, a set of normal samples is provided for reference. To tackle this problem, previous efforts attempt to construct various self-supervision tasks on those anomaly-free samples. These tasks include, but not limited to, sample reconstruction \cite{ssimae,fanogan,gn,memae,mnad,daad,pnet,cavga}, pseudo-outlier augmentation \cite{cutpaste,scadn,riad}, knowledge distillation\cite{mkd,us,stpm}, etc.

\begin{figure}[t]
  \centering
   \includegraphics[width=0.81\columnwidth]{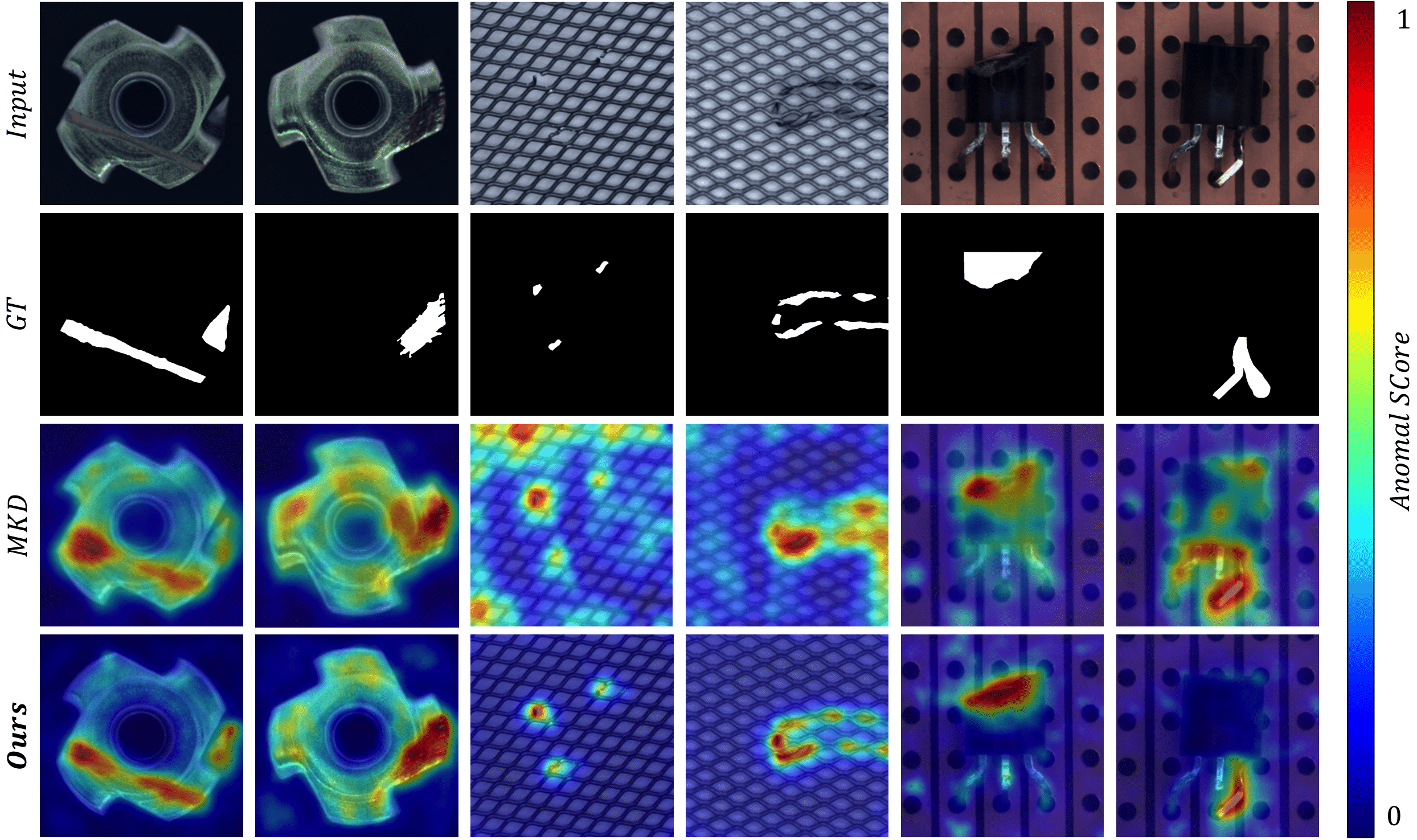}
   \caption{Anomaly detection examples on MVTec \cite{mvtec}. Multiresolution Knowledge Distillation (MKD) \cite{mkd} adopts the conventional KD architecture in Fig. \cref{fig:kd}(a). Our reverse distillation method is capable of precisely localising a variate of anomalies.} 
   \label{fig:mkd}
\end{figure}

In this study, we tackle the problem of unsupervised anomaly detection from the knowledge distillation-based point of view. In knowledge distillation (KD) \cite{kd,kdr}, knowledge is transferred within a teacher-student (T-S) pair. 
In the context of unsupervised AD, since the student experiences only normal samples during training, it is likely to generate discrepant representations from the teacher when a query is anomalous. This hypothesis forms the basis of KD-based methods for anomaly detection. However, this hypothesis is not always true in practice due to (1) the identical or similar architectures of the teacher and student networks (i.e., non-distinguishing filters \cite{mkd}) and (2) the same data flow in the T-S model during knowledge transfer/distillation. Though the use of a smaller student network partially addresses this issue
\cite{mkd,stpm}, the weaker representation capability of shallow architectures hinders the model from precisely detecting and localizing anomalies.

\begin{figure}[t]
  \centering
   \includegraphics[width=\columnwidth]{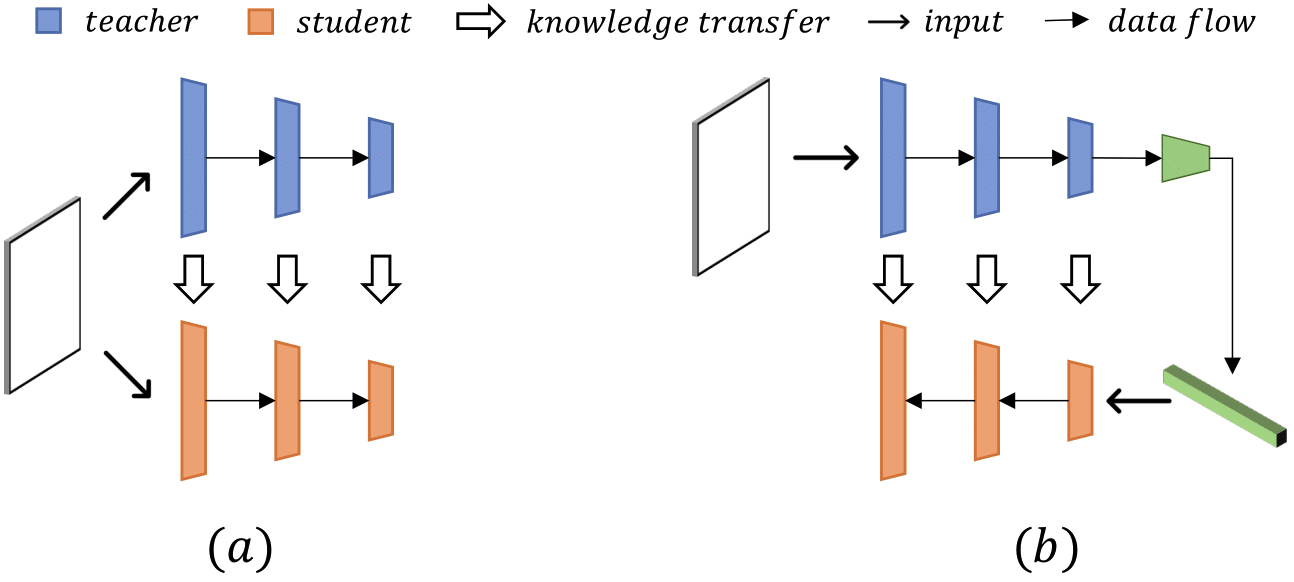}
   \caption{T-S models and data flow in (a) conventional KD framework \cite{kdr,mkd} and (b) our \emph{Reverse Distillation} paradigm.} 
   \label{fig:kd}
\end{figure}

To holistically address the issue mentioned above, we propose a new paradigm of knowledge distillation, namely \emph{Reverse Distillation}, for anomaly detection. We use simple diagrams in \cref{fig:kd} to highlight the systematic difference between conventional knowledge distillation and the proposed reverse distillation. First, unlike the conventional knowledge distillation framework where both teacher and student adopt the encoder structure, the T-S model in our reverse distillation consists of heterogeneous architectures: a teacher encoder and a student decoder. Second, instead of directly feeding the raw data to the T-S model simultaneously, the student decoder takes the low-dimensional embedding as input, targeting to mimic the teacher's behavior by restoring the teacher model's representations in different scales. From the regression perspective, our reverse distillation uses the student network to predict the representation of the teacher model. Therefore, "reverse" here indicates both the reverse shapes of teacher encoder and student decoder and the distinct knowledge distillation order where high-level representation is first distilled, followed by low-level features. It is noteworthy that our reverse distillation presents two significant advantages: $i)$ \emph{Non-similarity} structure. In the proposed T-S model, one can consider the teacher encoder as a down-sampling filter and the student decoder as an up-sampling filter. The "reverse structures" avoid the confusion caused by non-distinguishing filters \cite{mkd} as we discussed above. $ii)$ \emph{Compactness embedding}. The low-dimensional embedding fed to the student decoder acts as an information bottleneck for normal pattern restoration. Let's formulate anomaly features as perturbations on normal patterns. Then the compact embedding helps to prohibit the propagation of such unusual perturbations to the student model and thus boosts the T-S model's representation discrepancy on anomalies.   
Notably, traditional AE-based methods \cite{ssimae,memae,mnad,daad} detect anomalies utilising pixel differences, whereas we perform discrimination with dense descriptive features. Deep features as region-aware descriptors provide more effective discriminative information than per-pixel in images.

In addition, since the compactness of the bottleneck embedding is vital for anomaly detection (as discussed above), we introduce a one-class bottleneck embedding (OCBE) module to condense the feature codes further. Our OCBE module consists of a multi-scale feature fusion (MFF) block and one-class embedding (OCE) block, both jointly optimized with the student decoder. Notably, the former aggregates low- and high-level features to construct a rich embedding for normal pattern reconstruction. The latter targets to retain essential information favorable for the student to decode out the teacher's response.

We perform extensive experiments on public benchmarks. The experimental results indicate that our reverse distillation paradigm achieves comparable performance with prior arts. The proposed OCBE module further improves the performance to a new state-of-the-art (SOTA) record. Our main contributions are summarized as follows:
\begin{itemize}
  \item We introduce a simple, yet effective \emph{Reverse Distillation} paradigm for anomaly detection. 
  The encoder-decoder structure and reverse knowledge distillation strategy holistically address the non-distinguishing filter problem in conventional KD models, boosting the T-S model's discrimination capability on anomalies.
  \item We propose a \emph{one-class bottleneck embedding module} to project the teacher's high-dimensional features to a compact one-class embedding space. This innovation facilitates retaining rich yet compact codes for anomaly-free representation restoration at the student.
  \item We perform extensive experiments and show that our approach achieves new SOTA performance.
\end{itemize}


\section{Related Work}\label{sec:rw}
This section briefly reviews previous efforts on unsupervised anomaly detection. We will highlight the similarity and difference between the proposed method and prior arts.

Classical anomaly detection methods focus on defining a compact closed one-class distribution using normal support vectors. The pioneer studies include one-class support vector machine (OC-SVM) \cite{ocsvm} and support vector data description (SVDD) \cite{svdd}. To cope with high-dimensional data, DeepSVDD \cite{dsvdd} and PatchSVDD  \cite{psvdd} estimate data representations through deep networks. 

Another unsupervised AD prototype is the use of generative models, such as AutoEncoder (AE) \cite{vae} and Generative Adversarial Nets (GAN) \cite{gan}, for sample reconstruction. These methods rely on the hypothesis that generative models trained on normal samples only can successfully reconstruct anomaly-free regions, but fail for anomalous regions \cite{ssimae,fanogan,gn}. However, recent studies show that deep models generalize so well that even anomalous regions can be well-restored \cite{riad}. To address this issue, memory mechanism \cite{memae,mnad,daad} , image masking strategy \cite{scadn,riad} and pseudo-anomaly \cite{oig,g2d} are incorporated in reconstruction-based methods. However, these methods still lack a strong discriminating ability for real-world anomaly detection \cite{ssimae, mvtec}. Recently, Metaformer (MF) \cite{mf} proposes the use of meta-learning \cite{meta} to bridge model adaptation and reconstruction gap for reconstruction-based approaches. Notably, the proposed reverse knowledge distillation also adopts the encoder-decoder architecture, but it differs from construction-based methods in two-folds. First, the encoder in a generative model is jointly trained with the decoder, while our reverse distillation freezes a pre-trained model as the teacher. Second, instead of pixel-level reconstruction error, it performs anomaly detection on the semantic feature space.

Data augmentation strategy is also widely used. By adding pseudo anomalies in the provided anomaly-free samples, the unsupervised task is converted to a supervised learning task \cite{cutpaste,scadn,riad}. However, these approaches are prone to bias towards pseudo outliers and fail to detect a large variety of anomaly types. For example, CutPaste \cite{cutpaste} generates pseudo outliers by adding small patches onto normal images and trains a model to detect these anomalous regions. Since the model focuses on detecting local features such as edge discontinuity and texture perturbations, it fails to detect and localize large defects and global structural anomalies as shown in \cref{fig:large_ad}.

Recently, networks pre-trained on the large dataset are proven to be capable of extracting discriminative features for anomaly detection \cite{pretrain,spade,padim,panda,cutpaste,GAD}. With a pre-trained model, memorizing its anomaly-free features helps to identify anomalous samples \cite{spade,panda}. The studies in \cite{padim,GAD} show that using the Mahalanobis distance to measure the similarity between anomalies and anomaly-free features leads to accurate anomaly detection. Since these methods require memorizing all features from training samples, they are computationally expensive. 

Knowledge distillation from pre-trained models is another potential solution to anomaly detection. In the context of unsupervised AD, since the student model is exposed to anomaly-free samples in knowledge distillation, the T-S model is expected to generate discrepant features on anomalies in inference \cite{mkd,us,stpm}. To further increase the discrimnating capability of the T-S model on various types of abnormalities, different strategies are introduced. For instance, in order to capture multi-scale anomaly, US \cite{us} ensembles several models trained on normal data at different scales, and MKD \cite{mkd} propose to use multi-level features alignment. 
It should be noted that though the proposed method is also based on knowledge distillation, our reverse distillation is the first to adopt an encoder and a decoder to construct the T-S model. The heterogeneity of the teacher and student networks and reverse data flow in knowledge distillation distinguishes our method from prior arts. 
\section{Our Approach}
\label{sec:method}

\begin{figure*}[t]
  \centering
   \includegraphics[scale=0.53]{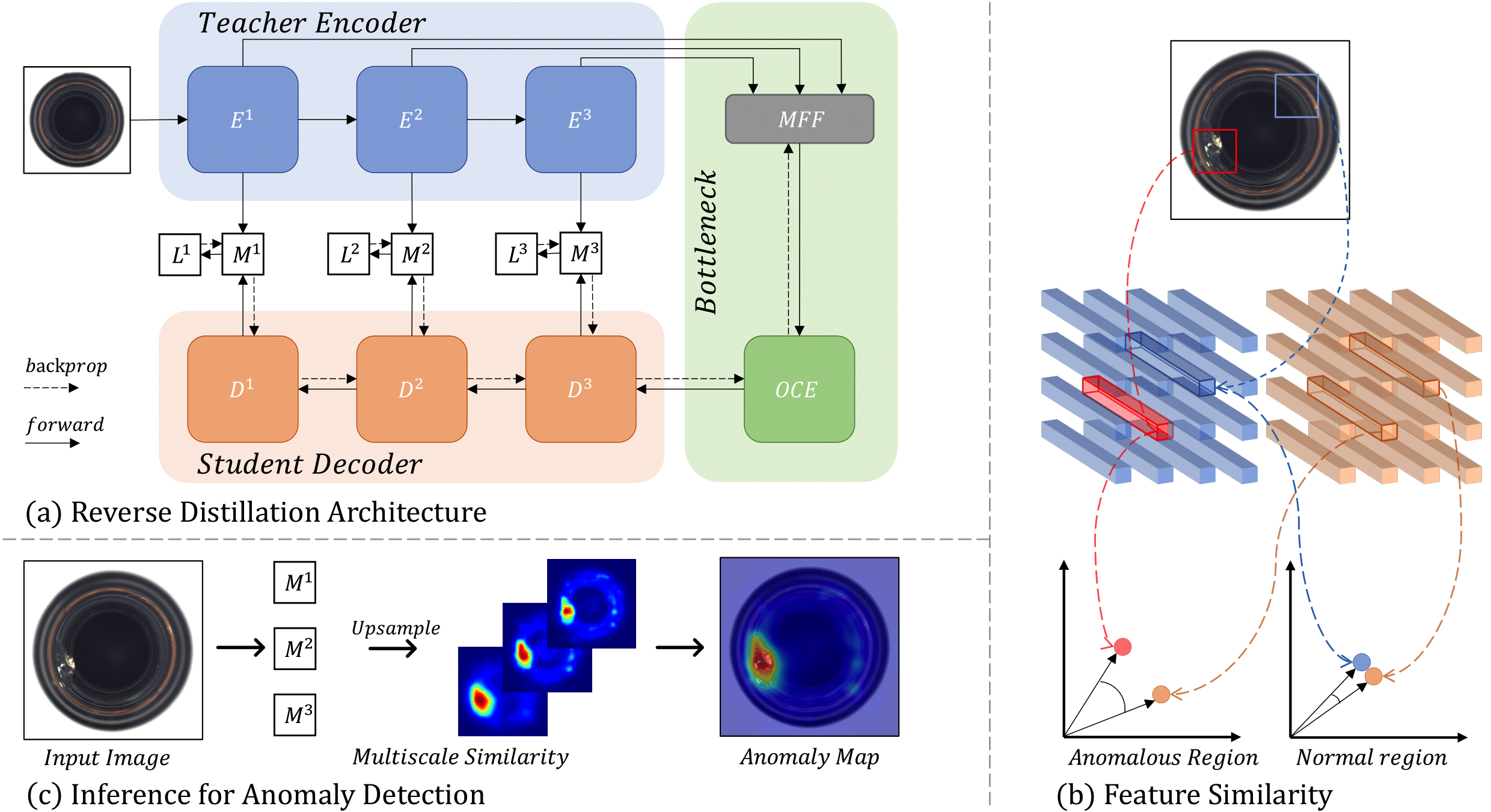}

   \caption{Overview of our reverse distillation framework for anomaly detection and localization. (a) Our model consists of a pre-trained teacher encoder $E$, a trainable one-class bottleneck embedding module (OCBE), and a student decoder $D$. We use a multi-scale feature fusion (MFF) block to ensemble low- and high-level features from $E$ and map them onto a compact code by one-class embedding (OCE) block. During training, the student $D$ learns to mimic the behavior of $E$ by minimizing the similarity loss $\mathcal{L}$. (b) During inference, $E$ extracts the features truthfully, while $D$ outputs anomaly-free ones. A low similarity between the feature vectors at the corresponding position of $E$ and $D$ implies an abnormality. (c) The final prediction is calculated by the accumulation of multi-scale similarity maps $M$.}
   \label{fig:pipline}
\end{figure*}

\textbf{Problem formulation:} Let $\mathcal{I}^t = \{I_1^t,..., I_n^t\}$ be a set of available anomaly-free images and $\mathcal{I}^q = \{I_1^q,..., I_m^q\}$ be a query set containing both normal and abnormal samples. The goal is to train a model to recognize and localize anomalies in the query set. In the anomaly detection setting, normal samples in both $\mathcal{I}^t$ and $\mathcal{I}^q$ follow the same distribution. Out-of-distribution samples are considered anomalies.

\textbf{System overview:} \cref{fig:pipline} depicts the proposed reserve distillation framework for anomaly detection. Our reverse distillation framework consists of three modules: a fixed pre-trained teacher encoder $E$, a trainable one-class bottleneck embedding module, and a student decoder $D$. Given an input sample $I\in \mathcal{I}^t$, the teacher $E$ extracts multi-scale representations. We propose to train a student $D$ to restore the features from the bottleneck embedding. 
During testing/inference, the representation extracted by the teacher $E$ can capture abnormal, out-of-distribution features in anomalous samples. However, the student decoder $D$ fails to reconstruct these anomalous features from the corresponding embedding. The low similarity of anomalous representations in the proposed T-S model indicates a high abnormality score. We argue that the heterogeneous encoder and decoder structures and reverse knowledge distillation order contribute a lot to the discrepant representations of anomalies. In addition, the trainable OCBE module further condenses the multi-scale patterns into an extreme low-dimensional space for downstream normal representation reconstruction. This further improves feature discrepancy on anomalies in our T-S model, as abnormal representations generated by the teacher model are likely to be abandoned by OCBE.   
In the rest of this section, we first specify the reverse distillation paradigm. Then, we elaborate on the OCBE module. Finally, we describe anomaly detection and localization using reserve distillation.

\subsection{Reverse Distillation}
\label{subsec:rd}

In conventional KD, the student network adopts a similar or identical neural network to the teacher model, accepts raw data/images as input, and targets to match its feature activations to the teacher's \cite{mkd,us}. In the context of one-class distillation for unsupervised AD, the student model is expected to generate highly different representations from the teacher when the queries are anomalous samples \cite{memae,mnad}. However, the activation discrepancy on anomalies vanishes sometimes, leading to anomaly detection failure. We argue that this issue is attributed to the similar architectures of the teacher and student nets and the same data flow during T-S knowledge transfer. To improve the T-S model's representation diversity on unknown, out-of-distribution samples, we propose a novel reserves distillation paradigm, where the T-S model adopts the encoder-decoder architecture and knowledge is distilled from teacher's deep layers to its early layers, i.e., high-level, semantic knowledge being transferred to the student first. To further facilitate the one-class distillation, we designed a trainable OCEB module to connect the teacher and student models (\cref{subsec:oce}).

In the reverse distillation paradigm, the teacher encoder $E$ aims to extract comprehensive representations. We follow previous work and use a pre-trained encoder on ImageNet \cite{imagenet} as our backbone $E$. To avoid the T-S model converging to trivial solutions, all parameters of teacher $E$ are frozen during knowledge distillation. We show in the ablation study that both ResNet \cite{He_2016_CVPR} and WideResNet \cite{BMVC2016_87} are good candidates, as they are capable of extracting rich features from images \cite{panda, padim, us, cutpaste}. 

To match the intermediate representations of $E$, the architecture of student decoder $D$ is symmetrical but reversed compared to $E$. The reverse design facilitates eliminating the response of the student network to abnormalities, while the symmetry allows it to have the same representation dimension as the teacher network. 
For instance, when we take ResNet as the teacher model, the student $D$ consists of several residual-like decoding blocks for mirror symmetry. Specifically, the downsampling in ResNet is realized by a convolutional layer with a kernel size of 1 and a stride of 2 \cite{He_2016_CVPR}. The corresponding decoding block in the student $D$ adopts deconvolutional layers \cite{deconv} with a kernel size of 2 and a stride of 2. More details on the student decoder design are given in \emph{Supplementary Material}.

In our reverse distillation, the student decoder $D$ targets to mimic the behavior of the teacher encoder $E$ during training. In this work, we explore multi-scale feature-based distillation for anomaly detection. The motivation behind this is that shallow layers of a neural network extract local descriptors for low-level information (e.g., color, edge, texture, etc.), while deep layers have wider receptive fields, capable of characterizing regional/global semantic and structural information. That is, low similarity of low- and high-level features in the T-S model suggests local abnormalities and regional/global structural outliers, respectively. 

Mathematically, let $\phi$ indicates the projection from raw data $I$ to the one-class bottleneck embedding space, the paired activation correspondence in our T-S model is $\{f^k_E=E^k(I), f^k_D=D^{k}(\phi)\}$, where $E^k$ and $D^k$ represent the $k^{th}$ encoding and decoding block in the teacher and student model, respectively. $f^k_E, f^k_D \in \mathbb{R}^{C_k\times H_k\times W_k}$, where $C_k$, $H_k$ and $W_k$ denote the number of channels, height and width of the $k^{th}$ layer activation tensor. For knowledge transfer in the T-S model, the cosine similarity is taken as the KD loss, as it is more precisely to capture the relation in both high- and low-dimensional information \cite{Zhu_2021_CVPR,Tung_2019_ICCV}. Specifically, for feature tensors $f^k_E$ and $f^k_D$, we calculate their vector-wise cosine similarity loss along the channel axis and obtain a 2-D anomaly map $M^k \in \mathbb{R}^{H_k\times W_k}$:
\begin{equation}
  M^k(h,w) = 1-\frac{(f^k_E(h,w))^T \cdot f^k_D(h,w)}{\left \| f^k_E(h,w) \right \| \left \| f^k_D(h,w) \right \|}
  \label{eq:hw_loss}.
\end{equation}
A large value in $M^k$ indicates high anomaly in that location. Considering the multi-scale knowledge distillation, the scalar loss function for student's optimization is obtained by accumulating multi-scale anomaly maps:
\begin{equation}
  \mathcal{L_{KD}} = \sum_{k=1}^{K} \bigg\{\frac{1}{H_kW_k}\sum_{h=1}^{H_k}\sum_{w=1}^{W_k}M^k(h,w)\bigg\}
  \label{eq:loss},
\end{equation}
where $K$ indicates the number of feature layers used in the experiment.

\subsection{One-Class Bottleneck Embedding}
\label{subsec:oce}

Since the student model $D$ attempts to restore representations of a teacher model $E$ in our reverse knowledge distillation paradigm, one can directly feed the activation output of the last encoding block in backbone to $D$. However, this naive connection has two shortfalls. First, the teacher model in KD usually has a high capacity. Though the high-capacity model helps extract rich features, the obtained high-dimensional descriptors likely have a considerable redundancy. The high freedom and redundancy of representations are harmful to the student model to decode the essential anomaly-free features. Second, the activation of the last encoder block in backbone usually characterizes semantic and structural information of the input data. Due to the reverse order of knowledge distillation, directly feeding this high-level representation to the student decoder set a challenge for low-level features reconstruction. Previous efforts on data reconstruction usually introduce skip paths to connect the encoder and decoder. However, this approach doesn't work in knowledge distillation, as the skip paths leak anomaly information to the student during inference.

To tackle the first shortfall above in one-class distillation, we introduce a trainable one-class embedding block to project the teacher model's high-dimensional representations into a low-dimensional space. Let's formulate anomaly features as perturbations on normal patterns. Then the compact embedding acts as an information bottleneck and helps to prohibit the propagation of unusual perturbations to the student model, therefore boosting the T-S model's representation discrepancy on anomalies. In this study, we adopt the 4th residule block of ResNet \cite{He_2016_CVPR} as the one-class embedding block.

To address the problem on low-level feature restoration at decoder $D$, the MFF block concatenates multi-scale representations before one-class embedding. To achieve representation alignment in feature concatenation, we down-sample the shallow features through one or more $3 \times 3$ convolutional layers with stride of 2, followed by batch normalization and ReLU activation function. Then a $1 \times 1$ convolutional layer with stride of 1 and a batch normalization with relu activation are exploited for a rich yet compact feature.

\begin{figure}[t]
  \centering
   \includegraphics[width=\columnwidth]{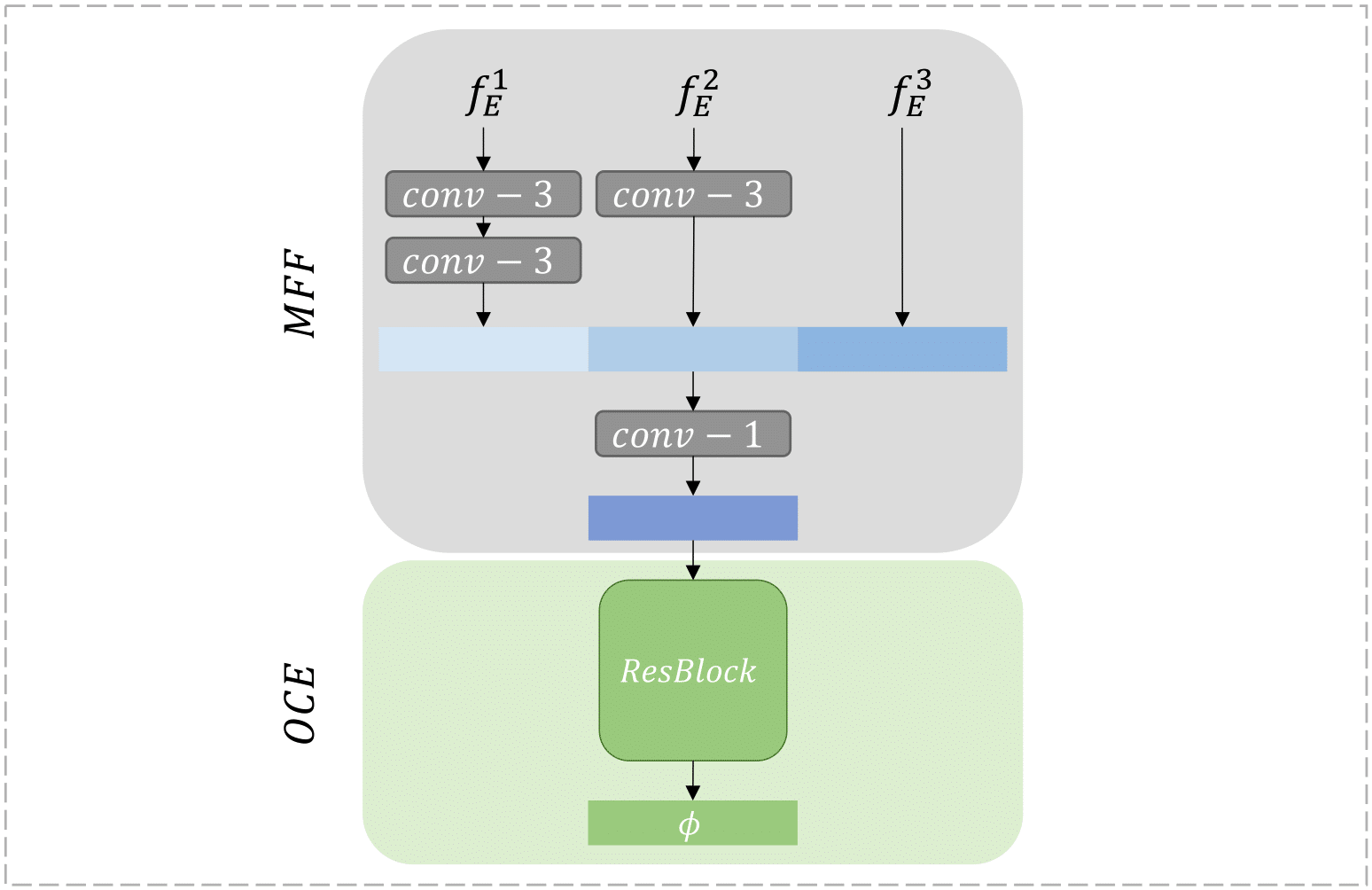}

   \caption{Our one-class bottleneck embedding module consists of trainable MFF and OCE blocks. MFF aligns multi-scale features from teacher $E$ and OCE condenses the obtained rich feature to a compact bottleneck code $\phi$.} 
   \label{fig:oce}
\end{figure}

We depict the OCBE module in \cref{fig:oce}, where MFF aggregates low- and high-level features to construct a rich embedding for normal pattern reconstruction and OCE targets to retain essential information favorable for the student to decode out the teacher's response. The convolutional layers in grey and ResBlock in green in \cref{fig:oce} are trainable and optimized jointly with the student model $D$ during knowledge distillation on normal samples.

\begin{table*}[!ht]
\footnotesize
\centering
\begin{tabular}{clccccccccccc}
\hline
\multicolumn{2}{c|}{Image Size} & \multicolumn{2}{c|}{128} & \multicolumn{9}{c}{256} \\ \hline
\multicolumn{2}{c|}{Category/Method} & \scriptsize {MKD\cite{mkd}} & \multicolumn{1}{c|}{Ours} & \scriptsize {GT\cite{gt}} & \scriptsize {GN\cite{gn}} & \scriptsize {US\cite{us}} & \scriptsize {PSVDD\cite{psvdd}} & \scriptsize {DAAD\cite{daad}} & \scriptsize {MF\cite{mf}} & \scriptsize {PaDiM\cite{padim}} & \scriptsize {CutPaste\cite{cutpaste}} & Ours \\ \hline
\multicolumn{1}{c|}{\multirow{6}{*}{\begin{turn}{90}Textures\end{turn}}} & \multicolumn{1}{l|}{Carpet} & 79.3 & \multicolumn{1}{c|}{99.2} & 43.7 & 69.9 & 91.6 & 92.9 & 86.6 & 94.0 & \textbf{99.8} & 93.9 & \textbf{98.9} \\ 
\multicolumn{1}{c|}{} & \multicolumn{1}{l|}{Grid} & 78.0 & \multicolumn{1}{c|}{95.7} & 61.9 & 70.8 & 81.0 & 94.6 & 95.7 & 85.9 & 96.7 & \textbf{100} & \textbf{100} \\ 
\multicolumn{1}{c|}{} & \multicolumn{1}{l|}{Leather} & 95.1 & \multicolumn{1}{c|}{100} & 84.1 & 84.2 & 88.2 & 90.9 & 86.2 & \textbf{99.2} & \textbf{100} & \textbf{100} & \textbf{100} \\ 
\multicolumn{1}{c|}{} & \multicolumn{1}{l|}{Tile} & 91.6 & \multicolumn{1}{c|}{99.4} & 41.7 & 79.4 & \textbf{99.1} & 97.8 & 88.2 & 99.0 & 98.1 & 94.6 & \textbf{99.3} \\ 
\multicolumn{1}{c|}{} & \multicolumn{1}{l|}{Wood} & 94.3 & \multicolumn{1}{c|}{98.8} & 61.1 & 83.4 & 97.7 & 96.5 & 98.2 & \textbf{99.2} & \textbf{99.2} & \textbf{99.1} & \textbf{99.2} \\ \cline{2-13} 
\multicolumn{1}{c|}{} & \multicolumn{1}{l|}{\textit{Average}} & \textit{87.7} & \multicolumn{1}{c|}{\textit{98.6}} & \textit{58.5} & \textit{77.5} & \textit{91.5} & \textit{94.5} & \textit{91.0} & \textit{95.5} & \textit{98.8} & \textit{97.5} & \textit{\textbf{99.5}} \\ \hline
\multicolumn{1}{c|}{\multirow{11}{*}{\begin{turn}{90}Objects\end{turn}}} & \multicolumn{1}{l|}{Bottle} & 99.4 & \multicolumn{1}{c|}{100} & 74.4 & 89.2 & 99.0 & 98.6 & 97.6 & 99.1 & \textbf{99.9} & 98.2 & \textbf{100} \\ 
\multicolumn{1}{c|}{} & \multicolumn{1}{l|}{Cable} & 89.2 & \multicolumn{1}{c|}{97.1} & 78.3 & 75.7 & 86.2 & 90.3 & 84.4 & \textbf{97.1} & 92.7 & 81.2 & \textbf{95.0} \\ 
\multicolumn{1}{c|}{} & \multicolumn{1}{l|}{Capsule} & 80.5 & \multicolumn{1}{c|}{89.5} & 67.0 & 73.2 & 86.1 & 76.7 & 76.7 & 87.5 & 91.3 & \textbf{98.2} & \textbf{96.3} \\
\multicolumn{1}{c|}{} & \multicolumn{1}{l|}{Hazelnut} & 98.4 & \multicolumn{1}{c|}{99.8} & 35.9 & 78.5 & 93.1 & 92.0 & 92.1 & \textbf{99.4} & 92.0 & 98.3 & \textbf{99.9} \\
\multicolumn{1}{c|}{} & \multicolumn{1}{l|}{Metal Nut} & 73.6 & \multicolumn{1}{c|}{99.2} & 81.3 & 70.0 & 82.0 & 94.0 & 75.8 & 96.2 & 98.7 & \textbf{99.9} & \textbf{100} \\
\multicolumn{1}{c|}{} & \multicolumn{1}{l|}{Pill} & 82.7 & \multicolumn{1}{c|}{93.3} & 63.0 & 74.3 & 87.9 & 86.1 & 90.0 & 90.1 & 93.3 & \textbf{94.9} & \textbf{96.6} \\ 
\multicolumn{1}{c|}{} & \multicolumn{1}{l|}{Screw} & 83.3 & \multicolumn{1}{c|}{91.1} & 50.0 & 74.6 & 54.9 & 81.3 & \textbf{98.7} & \textbf{97.5} & 85.8 & 88.7 & 97.0 \\ 
\multicolumn{1}{c|}{} & \multicolumn{1}{l|}{Toothbrush} & 92.2 & \multicolumn{1}{c|}{90.3} & 97.2 & 65.3 & 95.3 & \textbf{100} & 99.2 & \textbf{100} & 96.1 & 99.4 & \textbf{99.5} \\ 
\multicolumn{1}{c|}{} & \multicolumn{1}{l|}{Transistor} & 85.6 & \multicolumn{1}{c|}{99.5} & 86.9 & 79.2 & 81.8 & 91.5 & 87.6 & 94.4 & \textbf{97.4} & 96.1 & \textbf{96.7} \\ 
\multicolumn{1}{c|}{} & \multicolumn{1}{l|}{Zipper} & 93.2 & \multicolumn{1}{c|}{94.3} & 82.0 & 74.5 & 91.9 & 97.9 & 85.9 & \textbf{98.6} & 90.3 & \textbf{99.9} & 98.5 \\ \cline{2-13} 
\multicolumn{1}{c|}{} & \multicolumn{1}{l|}{\textit{Average}} & \textit{87.8} & \multicolumn{1}{c|}{\textit{95.4}} & \textit{71.6} & \textit{75.5} & \textit{85.8} & \textit{90.8} & \textit{88.8} & \textit{96.0} & \textit{93.8} & \textit{95.5} & \textit{\textbf{98.0}} \\ \hline
\multicolumn{2}{c}{\textit{Total Average}} & \textit{87.8} & \textit{96.5} & \textit{67.2} & \textit{76.2} & \textit{87.7} & \textit{92.1} & \textit{89.5} & \textit{95.8} & \textit{95.5} & \textit{96.1} & \textit{\textbf{98.5}} \\ \hline
\end{tabular}
\caption{\emph{Anomaly Detection} results on MVTec \cite{mvtec}. For each category with images of $256 \times 256$ resolution, methods achieved for the top two AUROC (\%) are highlighted in bold. Our method ranks first according to the average scores of {\bf textures}, {\bf objects} and overall.}
\label{tab:ad}
\end{table*}

\subsection{Anomaly Scoring}
At the inference stage, We first consider the measurement of pixel-level anomaly score for \emph{anomaly localization} (AL). When a query sample is anomalous, the teacher model is capable of reflecting abnormality in its features. However, the student model is likely to fail in abnormal feature restoration, since the student decoder only learns to restore anomaly-free representations from the compact one-class embedding in knowledge distillation. In other words, the student $D$ generates discrepant representations from the teacher when the query is anomalous. Following \cref{eq:hw_loss}, we obtain a set of anomaly maps from T-S representation pairs, where the value in a map $M_k$ reflects the point-wise anomaly of the $k^{th}$ feature tensors. To localize anomalies in a query image, we up-samples $M^k$ to image size. Let $\Psi$ denotes the bilinear up-sampling operation used in this study. Then a precise score map $S_{I^q}$ is formulated as the pixel-wise accumulation of all anomaly maps:
\begin{equation}
  S_{AL} = \sum_{i=1}^{L}\Psi(M^i)
  \label{eq:score}.
\end{equation}
In order to remove the noises in the score map, we smooth $S_{AL}$ by a Gaussian filter.

For \emph{anomaly detection}, averaging all values in a score map $S_{AL}$ is unfair for samples with small anomalous regions. The most responsive point exists for any size of anomalous region. Hence, we define the maximum value in $S_{AL}$ as sample-level anomaly score $S_{AD}$. The intuition is that no significant response exists in their anomaly score map for normal samples.


\section{Experiments and Discussions}
Empirical evaluations are carried on both the MVTec anomaly detection and localization benchmark and unsupervised one-class novelty detection datasets. In addition, we perform ablation study on the MVTec benchmark, investigating the effects of different modules/blocks on the final results.





\begin{table*}[!ht]
\footnotesize
\centering
\begin{tabular}{ccccccccccc}
\hline
\multicolumn{2}{c|}{Image Size} & \multicolumn{2}{c|}{128} & \multicolumn{7}{c}{256} \\ \hline
\multicolumn{2}{c|}{Category/Method} & MKD\cite{mkd} & \multicolumn{1}{c|}{Ours} & US\cite{us} & MF\cite{mf} & SPADE\cite{spade} & PaDiM\cite{padim} & RIAD\cite{riad} & CutPaste\cite{cutpaste} & Ours \\ \hline
\multicolumn{1}{c|}{\multirow{6}{*}{\begin{turn}{90}Textures\end{turn}}} & \multicolumn{1}{l|}{Carpet} & 95.6/- & \multicolumn{1}{c|}{98.1/95.3} & -/87.9 & -/87.8 & 97.5/94.7 & \textbf{99.1}/96.2 & 96.3/- & 98.3/- & 98.9/\textbf{97.0} \\
\multicolumn{1}{c|}{} & \multicolumn{1}{l|}{Grid} & 91.8/- & \multicolumn{1}{c|}{97.3/92.6} & -/95.2 & -/86.5 & 93.7/86.7 & 97.3/94.6 & 98.8/- & 97.5/- & \textbf{99.3}/\textbf{97.6} \\ 
\multicolumn{1}{c|}{} & \multicolumn{1}{l|}{Leather} & 98.1/- & \multicolumn{1}{c|}{99.0/98.6} & -/94.5 & -/95.9 & 97.6/97.2 & 99.2/97.8 & 99.4/- & \textbf{99.5}/- & 99.4/\textbf{99.1} \\  
\multicolumn{1}{c|}{} & \multicolumn{1}{l|}{Tile} & 82.8/- & \multicolumn{1}{c|}{92.6/84.8} & -/\textbf{94.6} & -/88.1 & 87.4/75.9 & 94.1/86.0 & 89.1/- & 90.5/- & \textbf{95.6}/90.6 \\ 
\multicolumn{1}{c|}{} & \multicolumn{1}{l|}{Wood} & 84.8/- & \multicolumn{1}{c|}{92.1/82.3} & -/\textbf{91.1} & -/84.8 & 88.5/87.4 & 94.9/\textbf{91.1} & 85.8/- & \textbf{95.5}/- & 95.3/90.9 \\ \cline{2-11} 
\multicolumn{1}{c|}{} & \multicolumn{1}{l|}{\textit{Average}} & \textit{90.6/-} & \multicolumn{1}{c|}{\textit{95.8/90.7}} & \textit{-/92.7} & \textit{-/88.6} & \textit{92.9/88.4} & \textit{96.9/93.2} & 93.9/- & \textit{96.3/-} & \textit{\textbf{97.7/95.0}} \\ \hline
\multicolumn{1}{c|}{\multirow{11}{*}{\begin{turn}{90}Objects\end{turn}}} & \multicolumn{1}{l|}{Bottle} & 96.3/- & \multicolumn{1}{c|}{98.2/94.7} & -/93.1 & -/88.8 & 98.4/95.5 & 98.3/94.8 & 98.4/- & 97.6/- & \textbf{98.7/96.6} \\ 
\multicolumn{1}{c|}{} & \multicolumn{1}{l|}{Cable} & 82.4/- & \multicolumn{1}{c|}{97.8/90.5} & -/81.8 & -/\textbf{93.7} & 97.2/90.9 & 96.7/88.8 & 84.2/- & 90.0/- & \textbf{97.4}/91.0 \\ 
\multicolumn{1}{c|}{} & \multicolumn{1}{l|}{Capsule} & 95.9/- & \multicolumn{1}{c|}{96.5/87.2} & -/\textbf{96.8} & -/87.9 & \textbf{99.0}/93.7 & 98.5/93.5 & 92.8/- & 97.4/- & 98.7/95.8 \\
\multicolumn{1}{c|}{} & \multicolumn{1}{l|}{Hazelnut} & 94.6/- & \multicolumn{1}{c|}{98.8/89.2} & -/\textbf{96.5} & -/88.6 & \textbf{99.1}/95.4 & 98.2/92.6 & 96.1/- & 97.3/- & 98.9/95.5 \\ 
\multicolumn{1}{c|}{} & \multicolumn{1}{l|}{Metal Nut} & 86.4/- & \multicolumn{1}{c|}{96.6/84.1} & -/94.2 & -/86.9 & \textbf{98.1/94.4} & 97.2/85.6 & 92.5/- & 93.1/- & 97.3/92.3 \\ 
\multicolumn{1}{c|}{} & \multicolumn{1}{l|}{Pill} & 89.6/- & \multicolumn{1}{c|}{97.0/90.0} & -/96.1 & -/93.0 & 96.5/94.6 & 95.7/92.7 & 95.7/- & 95.7/- & \textbf{98.2/96.4} \\ 
\multicolumn{1}{c|}{} & \multicolumn{1}{l|}{Screw} & 96.0/- & \multicolumn{1}{c|}{98.3/94.4} & -/94.2 & -/95.4 & 98.9/96.0 & 98.5/94.4 & 98.8/- & 96.7/- & \textbf{99.6/98.2} \\ 
\multicolumn{1}{c|}{} & \multicolumn{1}{l|}{Toothbrush} & 96.1/- & \multicolumn{1}{c|}{98.2/86.7} & -/93.3 & -/87.7 & 97.9/93.5 & 98.8/93.1 & 98.9/- & 98.1/- & \textbf{99.1/94.5} \\ 
\multicolumn{1}{c|}{} & \multicolumn{1}{l|}{Transistor} & 76.5/- & \multicolumn{1}{c|}{97.6/85.2} & -/66.6 & -/\textbf{92.6} & 94.1/87.4 & \textbf{97.5}/84.5 & 87.7/- & 93.0/- & 92.5/78.0 \\ 
\multicolumn{1}{c|}{} & \multicolumn{1}{l|}{Zipper} & 93.9/- & \multicolumn{1}{c|}{97.0/92.3} & -/95.1 & -/93.6 & 96.5/92.6 & \textbf{98.5/95.9} & 97.8/- & 99.3/- & 98.2/95.4 \\ \cline{2-11} 
\multicolumn{1}{c|}{} & \multicolumn{1}{l|}{\textit{Average}} & \textit{90.8/-} & \multicolumn{1}{c|}{\textit{97.6/89.4}} & \textit{-/90.8} & \textit{-/90.8} & \textit{97.6/\textbf{93.4}} & \textit{97.8/91.6} & 94.3/- & \textit{95.8/-} & \textit{\textbf{97.9/93.4}} \\ \hline
\multicolumn{2}{c}{\textit{Total Average}} & \textit{90.7/-} & \textit{97.0/89.9} & \textit{-/91.4} & \textit{-/90.1} & \textit{96.5/91.7} & \textit{97.5/92.1} & 94.2/- & \textit{96.0-} & \textit{\textbf{97.8/93.9}} \\ \hline
\end{tabular}
\caption{\emph{Anomaly Localization} results with AUROC and PRO on \textbf{MVTec} \cite{mvtec}. AUROC represents a pixel-wise comparison, while PRO focuses on region-based behavior. We show the best results for AUROC and PRO in bold. Remarkable, our approach is robust and represents state-of-the-art performance under both metrics.}
\label{tab:al}
\end{table*}

\subsection{Anomaly Detection and Localization}
\textbf{Dataset.} {\bf MVTec} \cite{mvtec} contains 15 real-world datasets for \emph{anomaly detection}, with 5 classes of {\bf textures} and 10 classes of {\bf objects}. The training set comprises a total of 3,629 anomaly-free images. The test set has both anomalous and anomaly-free images, totaling 1,725. Each class has multiple defects for testing. In addition, pixel-level annotations are available in the test dataset for \emph{anomaly localization} evaluation.

\textbf{Experimental settings.} All images in MVTec are resized to a specific resolution (e.g. $128 \times 128$, $256 \times 256$ etc.). Following convention in prior works, anomaly detection and localization are performed on one category at a time. 
In this experiment, we adopt WideResNet50 as Backbone $E$ in our T-S model. We also report the AD results with ResNet18 and ResNet50 in ablation study. To train our reserve distillation model, we utilize Adam optimizer \cite{adam} with $\beta = (0.5, 0.999)$. The learning rate is set to 0.005. We train 200 epochs with a batch size of 16. A Gaussian filter with $\sigma = 4$ is used to smooth the anomaly score map (as described in Sec. 3.3). 

For \emph{Anomaly dectction}, we take area under the receiver operating characteristic (AUROC) as the evaluation metric. We include prior arts in this experiments, including MKD\cite{mkd}, GT\cite{gt}, GANomaly (GN) \cite{gn}, Uninformed Student (US) \cite{us}, PSVDD \cite{psvdd}, DAAD \cite{daad}, MetaFormer (MF) \cite{mf}, PaDiM (WResNet50) \cite{padim} and CutPaste \cite{cutpaste}.

For \emph{Anomaly Localization}, we report both AUROC and per-region-overlap (PRO) \cite{us}. Different from AUROC, which is used for per-pixel measurement, the PRO score treats anomaly regions with any size equally. The comparison baselines includes MKD \cite{mkd}, US \cite{us}, MF \cite{mf}, SPADE (WResNet50) \cite{spade,panda}, PaDiM (WResNet50) \cite{padim}, RIAD \cite{riad} and CutPaste \cite{cutpaste}. 

\begin{figure}[ht]
  \centering
   \includegraphics[width=\columnwidth]{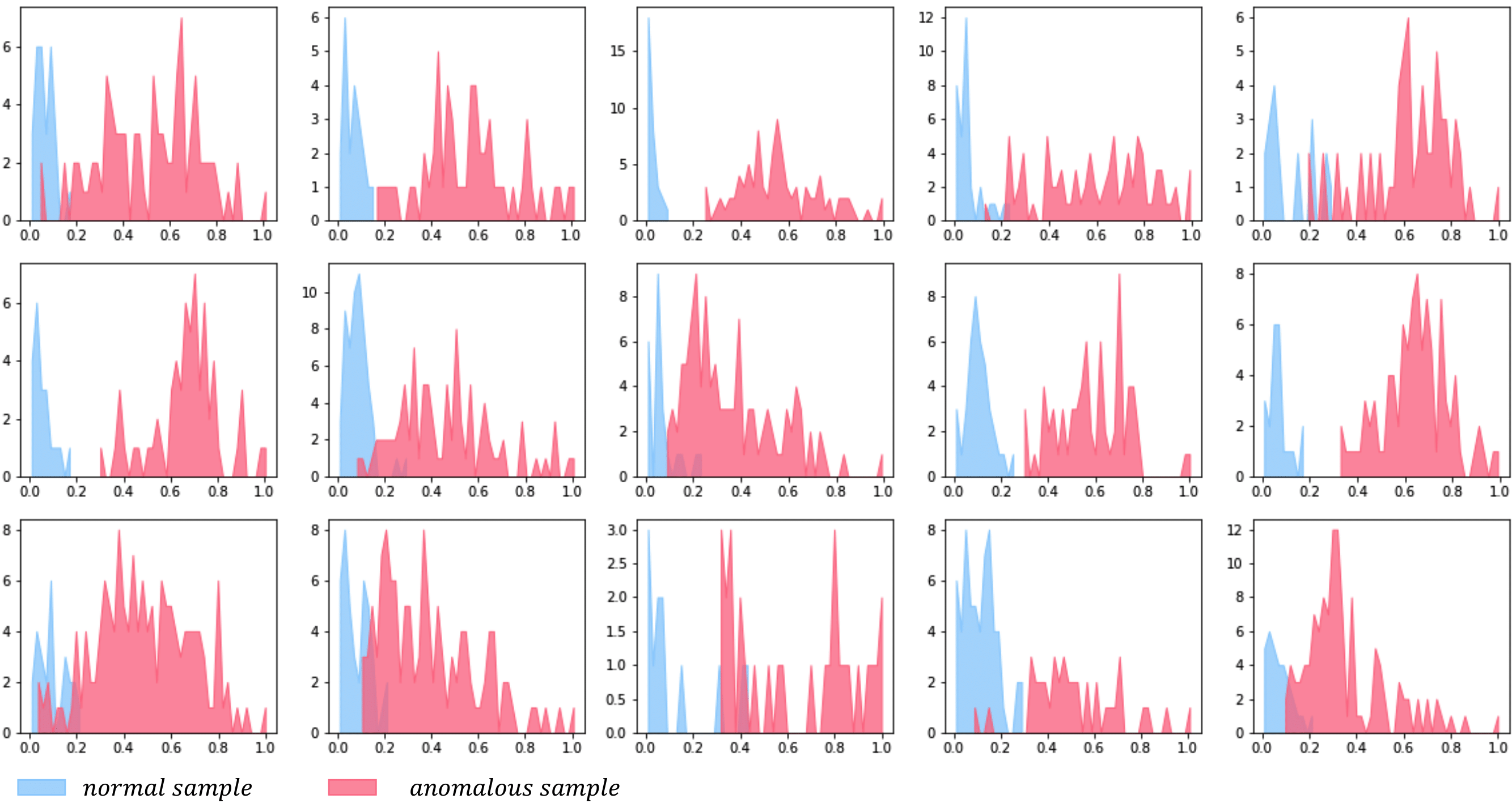}
   \caption{Histogram of anomaly scores for all categories of MVTec \cite{mvtec} (x-axis: anomaly score from 0 to 1, and y-axis: count).} 
   \label{fig:ad_vis}
\end{figure}

\textbf{Experimental results and discussions.}
Anomaly detection results on MVTec are shown in \cref{tab:ad}. The average outcome shows that our method exceeds SOTA by \textbf{2.5\%}. For {\bf textures} and {\bf objects}, our model reaches new SOTA of \textbf{99.5\%} and \textbf{98.0\%} of AUROC, respectively. The statistics of the anomaly scores are shown in \cref{fig:ad_vis}. The non-overlap distribution of normal (blue) and anomalies (red) indicates the strong AD capability in our T-S model.

\begin{figure}[ht]
\small
  \centering
   \includegraphics[width=\columnwidth]{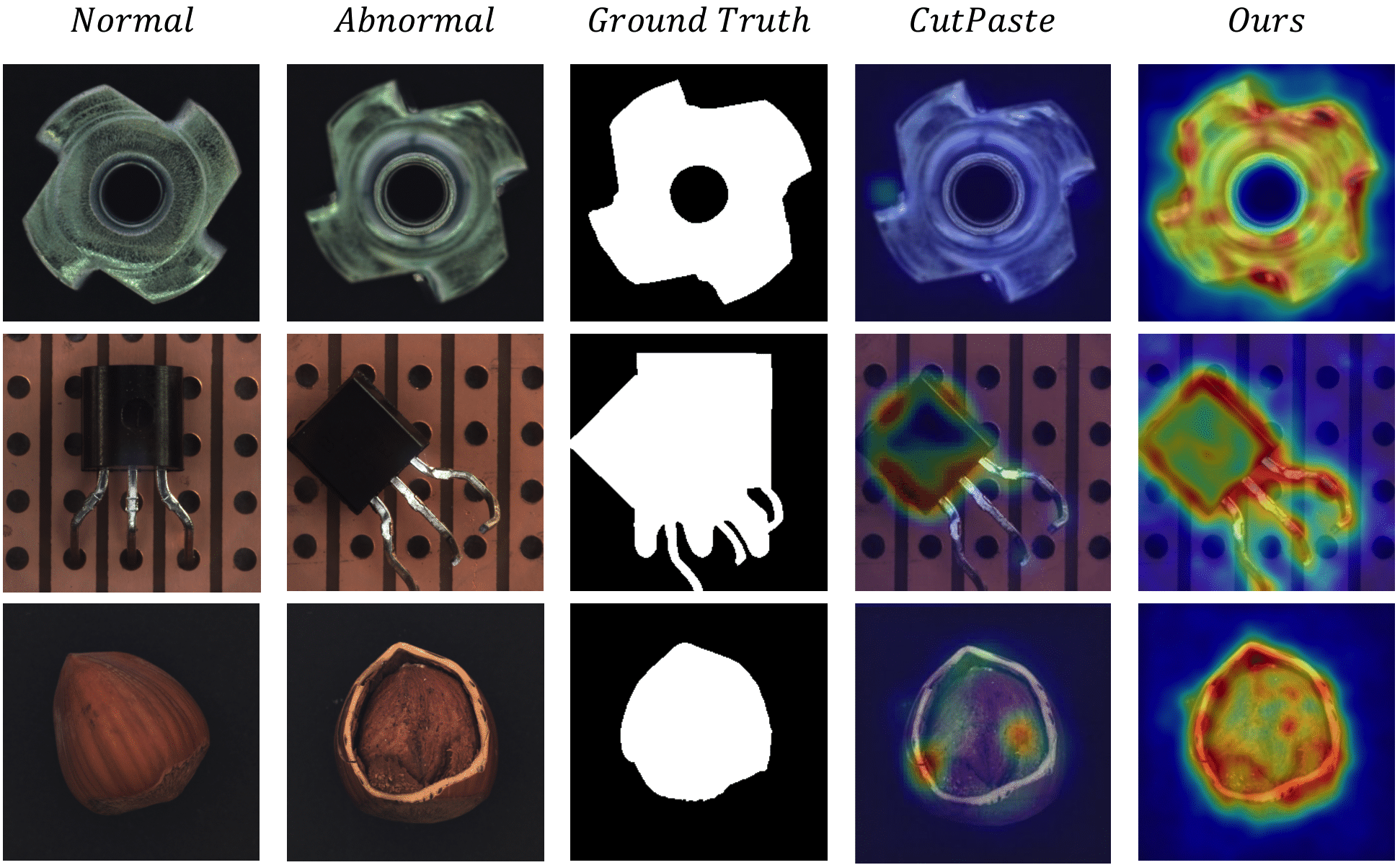}
   \caption{Anomalies from top to bottom: \emph{"flip"} on \emph{"metal nut"}, \emph{"misplaced"} on \emph{"transistor"} and \emph{"crack"} on \emph{"hazelnut"}. Normal  samples are provided as reference.} 
   \label{fig:large_ad}
\end{figure}

\begin{figure}[ht]
  \centering
   \includegraphics[width=\columnwidth]{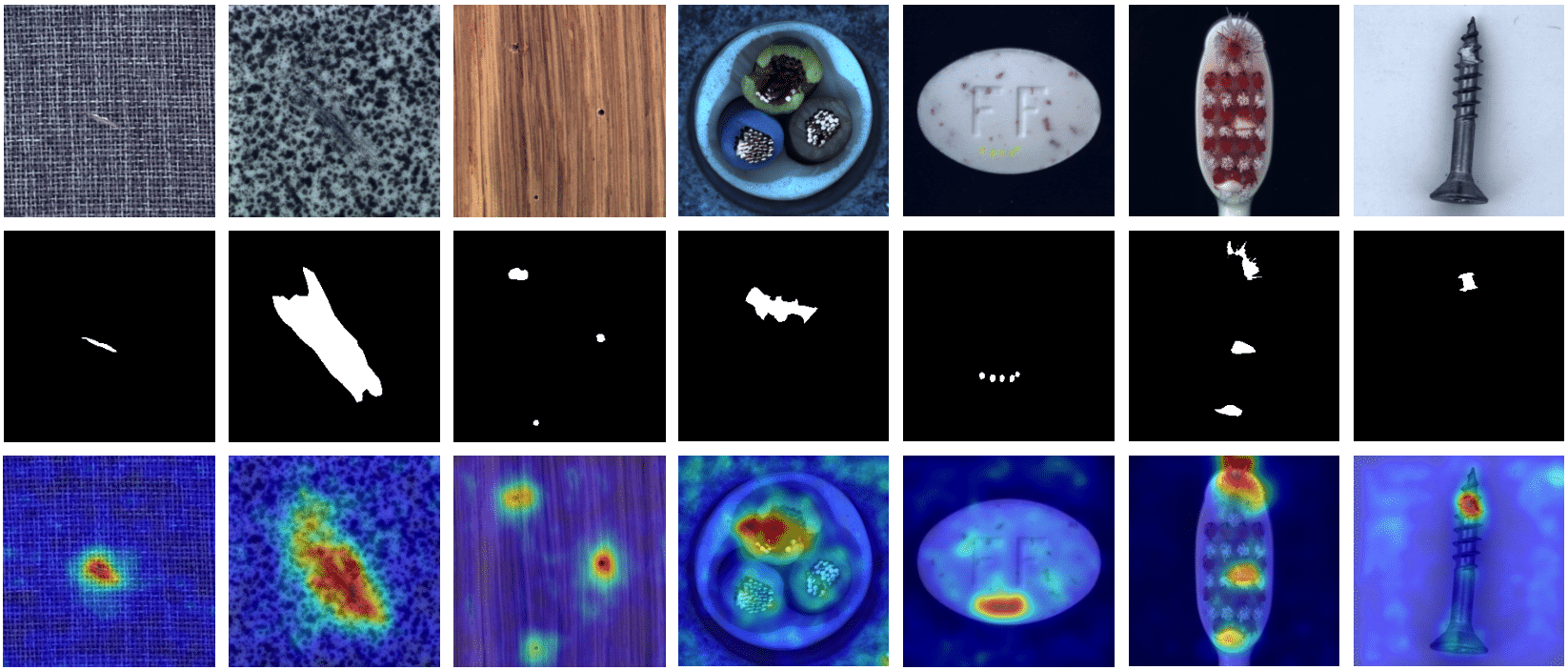}
   \caption{Visualization on tiny or inconspicuous anomalies. From left to right: \emph{carpet}, \emph{tile}, \emph{wood}, \emph{cable}, \emph{pill}, \emph{toothbrush}, and \emph{screw}.}
   \label{fig:small_ad}
\end{figure}

Quantitative results on anomaly localization are summarized in \cref{tab:al}. For both AUROC and PRO average scores over all categories, our approach surpasses state-of-the-art with \textbf{97.8\%} and \textbf{93.9\%}. 
To investigate the robustness of our method to various anomalies, we classify the defect types into two categories: large defects or structural anomalies and tiny or inconspicuous defects, and qualitative evaluate the performance by visualization in \cref{fig:large_ad} and \cref{fig:small_ad}. 
Compared to the runner-up (i.e. CutPaste \cite{cutpaste}) in \cref{tab:ad}, our method produces a significant response to the whole anomaly region. 

\textbf{Complexity analysis}. Recent pre-trained model based approaches achieve promising performance by extracting features from anomaly-free samples as a measurement\cite{padim,spade}. However, storing feature models leads to large memory consumption. In comparison, our approach achieves better performance depending only on an extra CNN model. As shown in \cref{tab:complexity}. Our model obtain performance gain with low time and memory complexity.
\begin{table}[ht]
\footnotesize
\centering
\begin{tabular}{c|c|c|c}
\hline
Methods & Infer. time & Memory & Performance \\ \hline
SPADE (WResNet50)    & 1.40                     & 1400                & 85.5/96.5/91.7     \\ \hline
PaDiM (WResNet50)        & 0.95                  & 3800             & 95.5/97.5/92.1          \\ \hline
\textbf{Ours (WResNet50)} & \textbf{0.31}         & \textbf{352}     & \textbf{98.5/97.8/93.9} \\ \hline
\end{tabular}
\caption{Comparison of pre-trained based approaches in terms of inference time (second on Intel i7), memory usage (MB), and performance (AD-AUROC/AL-AUROC/AL-PRO ) on MVTec \cite{mvtec}.}
\label{tab:complexity}
\end{table}

\textbf{Limitations.} We observe that the localization performance on the \emph{transistor} dataset is relatively weak, despite the good AD performance. 
This performance drop is caused by misinterpretation between prediction and annotation. As shown in \cref{fig:large_ad}, our method localize the misplaced regions, while the ground truth covers both misplaced and original areas. Alleviating this problem requires associating more features with contextual relationships. We empirically find that a higher-level feature layer with a wider perceptive field can improve the performance. For instance, anomaly detection with the second and third layer features achieves 94.5\% AUROC, while using only the third layer improve the performance to 97.3\%. 
In addition, reducing image resolution to $128 \times 128$ also achieves 97.6\% AUROC. We present more cases of anomaly detection and localization, both positive and negative, in the \emph{supplementary material}.

\subsection{One-Class Novelty Detection}
To evaluate the generality of proposed approach, we conduct \emph{one-class novelty detection} experiment on 3 sematic datasets \cite{salehi2021unified}, {\bf MNIST \cite{mnist}, FashionMNIST \cite{fmnist}} and {\bf CIFAR10 \cite{cifar10}}. MNIST is a hand-written digits dataset from numbers 0-9. FashionMNIST consists of images from 10 fashion product classes. Both datasets includes 60K samples for training and 10K samples for testing, all in resolution of $28 \times 28$.
CIFAR10 is a challenging dataset for novelty detection because of its inclusion of diverse natural objects. It includes 50K training images and 10K test images with scale of $32 \times 32$ in 10 categories. 

Following the protocol mentioned in \cite{ocgan}, we train the model with samples from a single class and detect novel samples. Note that the novelty score is defined as the sum of scores in the similarity map. The baselines in this experiment include LSA \cite{lsa}, OCGAN \cite{ocgan}, HRN \cite{hrn}, DAAD \cite{daad} and MKD \cite{mkd}. We also include the comparision with OiG \cite{oig} and G2D \cite{g2d} on {\bf Caltech-256 \cite{caltech}}.

\cref{tab:ocnd} summarizes the quantitative results on the three datasets. Remarkably, our approach produces excellent results. Details of the experiments and the results of per-class comparisons are provided in the \emph{Supplementary Material}.


\begin{table}
\footnotesize
\centering
\begin{tabular*}{\columnwidth}{@{}@{\extracolsep{\fill}}c|cccc@{}}
\hline
Method        & MNIST         & F-MNIST       & CIFAR10       & Caltech-256   \\ \hline
LSA\cite{lsa}           & 97.5          & 92.2          & 64.1          & -             \\ \hline
OCGAN\cite{ocgan}         & 97.3          & 87.8          & 65.7          & -             \\ \hline
HRN\cite{hrn}           & 97.6          & 92.8          & 71.3          & -             \\ \hline
DAAD\cite{daad}          & 99.0          & -             & 75.3          & -             \\ \hline
MKD\cite{mkd}           & 98.7          & 94.5          & 84.5          & -             \\ \hline
G2D\cite{g2d}           & -             & -             & -             & 95.7          \\ \hline
OiG\cite{oig}           & -             & -             & -             & 98.2          \\ \hline
\textbf{Ours} & \textbf{99.3} & \textbf{95.0} & \textbf{86.5} & \textbf{99.9} \\ \hline
\end{tabular*}
\caption{AUROC(\%) results for One-Class Novelty Detection. The best results are marked in bold.}
\label{tab:ocnd}
\end{table}

\subsection{Ablation analysis}
We investigate effective of OCE and MFF blocks on AD and reports the numerical results in \cref{tab:oce}. We take the pre-trained residual block \cite{He_2016_CVPR} as baseline. Embedding from pre-trained residual block may contain anomaly features, which decreases the T-S model's representation discrepancy. Our trainable OCE block condenses feature codes and the MFM block fuses rich features into embedding, allows for more accurate anomaly detection and localization.

\begin{table}[ht]
\footnotesize
\centering
\begin{tabular*}{\hsize}{@{}@{\extracolsep{\fill}}c|ccc@{}}
\hline
Metric & Pre & Pre+OCE & Pre+OCE+MFM \\ \hline
$AUROC_{AD}$ & 96.0 & 97.9 & \textbf{98.5} \\ \hline
$AUROC_{AL}$ & 96.9 & 97.4 & \textbf{97.8} \\ \hline
$RPO$ & 91.2 & 92.4 & \textbf{93.9} \\ \hline
\end{tabular*}
\caption{Ablation study on pre-trained bottleneck, OCE, and MFF.}
\label{tab:oce}
\end{table}

\cref{tab:bb} displays qualitative comparisons of different backbone networks as the teacher model. 
Intuitively, a deeper and wider network usually have a stronger representative capacity, which facilitates detecting anomalies precisely. Noteworthy that even with a smaller neural network such as ResNet18, our reverse distillation method still achieves excellent performance.    

\begin{table}[ht]
\footnotesize
\centering
\begin{tabular*}{\hsize}{@{}@{\extracolsep{\fill}}c|ccc@{}}
\hline
Backbone & ResNet18 & ResNet50 & WResNet50 \\ \hline
$AUROC_{AD}$ & 97.9 & 98.4 & \textbf{98.5} \\ \hline
$AUROC_{AL}$ & 97.1 & 97.7 & \textbf{97.8} \\ \hline
$RPO$ & 91.2 & 93.1 & \textbf{93.9} \\ \hline
\end{tabular*}
\caption{Quantitative comparison with different backbones.}
\label{tab:bb}
\end{table}

Besides, we also explored the impact of different network layers on anomaly detection and shown the results in \cref{tab:map}. For single-layer features, $M^2$ yields the best result as it trades off both local texture and global structure information. Multi-scale feature fusion helps to cover more types of anomalies.   

\begin{table}[ht]
\footnotesize
\centering
\begin{tabular*}{\hsize}{@{}@{\extracolsep{\fill}}c|ccccc@{}}
\hline
Score Map & $M^1$ & $M^2$ & $M^3$ & $M^{2,3}$ & $M^{1,2,3}$ \\ \hline
$AUROC_{AD}$ & 90.1 & 97.5 & 97.2 &98.0 & 98.5\\ \hline
$AUROC_{AL}$ & 94.0 & 96.9 & 96.9 &97.6 & 97.8\\ \hline
$RPO$ & 88.6 & 92.6 & 89.5 &93.2 & 93.9\\ \hline
\end{tabular*}
\caption{Ablation study on multi-scale feature distillation.}
\label{tab:map}
\end{table}

\section{Conclusion}

We proposed a novel knowledge distillation paradigm, reverse distillation, for anomaly detection. It holistically addressed the problem in previous KD-based AD methods and boosted the T-S model's response on anomalies. 
In addition, we introduced trainable one-class embedding and multi-scale feature fusion blocks in reverse distillation to improve one-class knowledge transfer. Experiments showed that our method significantly outperformed previous arts in anomaly detection, anomaly localization, and novelty detection.

{\small
\bibliographystyle{ieee_fullname}
\bibliography{egbib}
}

\end{document}